\documentclass[letterpaper, 10pt]{IEEEtran}
\IEEEoverridecommandlockouts
\usepackage[pdftex]{graphicx}

\graphicspath{{../pdf/}{../jpeg/}}
\DeclareGraphicsExtensions{.pdf,.jpeg,.png,.jpg}

\usepackage{amssymb} 
\usepackage{graphicx}
\usepackage[natural]{xcolor}

\usepackage{changes}

\usepackage{siunitx}

\usepackage[cmex10]{amsmath}
\usepackage[tight,footnotesize]{subfigure}

\usepackage{url}

\usepackage{colortbl}
\usepackage{tabulary}
\usepackage{etoolbox}
\usepackage{tabularx}
\usepackage{multirow}
\usepackage{float}

\begin{document}

\title{\LARGE \bf Coopetitive Soft Gating Ensemble }


\author{Stephan Deist, Maarten Bieshaar, Jens Schreiber,  Andr\'{e} Gensler, and Bernhard Sick\\
	  Intelligent Embedded Systems Group, University of Kassel (Germany)\\	
		email: \{stephan.deist, mbieshaar, jens.schreiber, gensler, bsick\}@uni-kassel.de
	%
}

\maketitle	
\begin{abstract}
    In this article, we propose the \textit{Coopetititve Soft Gating Ensemble} or \textit{CSGE} for general machine learning tasks and interwoven systems.
    The goal of machine learning is to create models that generalize well for unknown datasets. Often, however, the problems are too complex to be solved with a single model, so several models are combined. Similar, Autonomic Computing requires the integration of different systems. Here, especially, the local, temporal online evaluation and the resulting \mbox{(re-)weighting} scheme of the CSGE makes the approach highly applicable for self-improving system integrations.
    To achieve the best potential performance the \textit{CSGE} can be optimized according to arbitrary loss functions making it accessible for a broader range of problems. We introduce a novel training procedure including a hyper-parameter initialisation at its heart.
    We show that the \textit{CSGE} approach reaches state-of-the-art performance for both classification and regression tasks. Further on, the \textit{CSGE} provides a human-readable quantification on the influence of all base estimators employing the three weighting aspects. Moreover, we provide a scikit-learn compatible implementation.
\end{abstract}



\section{Introduction}
\label{sec:einleitung}

The primary goal of \textit{machine learning (ML)} is to create models from training data, which have a high generalization capability for unseen data.
Often the problems are so complex that one single \textit{estimator} cannot handle the whole scope.
These problems can, e.g., be tackled with a combination of multiple estimators instead of an individual estimator. This attempt of combining multiple estimators is called \textit{ensemble}. In many fields, ensembles can achieve state-of-the-art performance. Popular ensemble methods are \textit{Boosting} \cite{Schapire1990}, \textit{Bagging}, \cite{Breiman1996} or \textit{Stacking} \cite{Smyth1999}. \cite{hansen_salamon} shows that ensembles often lead to better results than using a single estimator. When considering the \textit{Bias–variance tradeoff}~\cite{KW96}, ensembles can reduce both variance and bias and therefore result in stronger models.

The combination of different models in an ensemble can be compared with the integration of different systems. Due to the common weighting of different models, the prediction is decisively determined by the mutual influence of individual models, as with interwoven systems~\cite{TomfordeRBW16}. In addition to this similarity, models are usually heterogeneous, e.g., linear and non-linear models. Ultimately, similar to interwoven system predictions are linked to uncertainty ~\cite{TomfordeRBW16}. Recently, an ensemble method called \textit{Coopetitive Soft Gating Ensemble} or \textit{CSGE} was proposed for wind power forecasts. However, CSGE also offers a basic technology to make the functionality of \textit{self-improving system integration (SISSY)} more robust against interference and to better avoid uncertainty by its weighting and optimization scheme \cite{BellmanTW14a}.

In~\cite{GS16},~\cite{GS18}, and \cite{GS17} it is statistically shown that the CSGE can achieve state-of-the-art performance in the area of power forecasting. In this article, we aim to extend the original approach to general \textit{ML} problems to show its potential for a wide range of problems including SISSY applications.
The idea of the CSGE is to gradually weight individual ensemble members according to their historically observed performance of different aspects. In particular, there are three aspects which take influence on the weight: First, the overall performance of the estimator. Second, the local performance of the estimator in similar historical situations. Third, time-dependent effects modeling the autocorrelation in the estimator's outcome. Aspect two and three are optimized w.r.t (current) online data, and the first aspect is determined based on the training data beforehand.

Hence, the ability to assess the individual performance of each base estimator for those three aspects can be interpreted, 
regarding Organic Computing (OC)~\cite{MSSU11} and Autonomic Computing (AC)~\cite{kephart2003vision}, as self-awareness and self-improving capabilities for SISSY.

\section{Main Contribution}
\label{sec:Ziele}

The main contribution of this article is an extended coopetitive soft gating ensemble approach.
It generalizes the original CSGE method proposed in~\cite{GS16, GS18} for wind power forecasting to other \textit{ML} tasks including regression, classification, and time series forecasting. The main contributions of this article are: 

\begin{itemize}
    \item The loss function of the \textit{CSGE} can be chosen by the user with only minimal constraints allowing optimization of arbitrary loss functions to make it available for SISSY and \textit{ML} applications.
    \item A novel heuristic to choose the hyper-parameters of the \textit{CSGE} training algorithm is reducing the required number of adjustable parameters.
    \item An extensive evaluation of our approach on common real-world reference datasets, in which we show that our \textit{CSGE} approach reaches state-of-the-art performance compared to other ensembles methods. Additionally, the \textit{CSGE} allows quantifying the influence of all base estimators utilizing the three weighting aspects in a human-readable way.
    \item A scikit-learn compatible implementation of the \textit{CSGE}\footnote{\url{https://git.ies.uni-kassel.de/csge/csge}}.
\end{itemize}

The remainder of this article is structured as follows. In Section~\ref{sec:related_work}, we review the related work in the field of ensemble method for \textit{ML}. Afterward, in Section~\ref{sec:method}, we introduce our \textit{CSGE} approach. In Section~\ref{sec:experimente}, we present the evaluation of our \textit{CSGE} on three synthetic datasets, four reference classification, and real-world regression datasets. Therefore, showing its applicability to a wide range of problems. Finally, in Section~\ref{sec:conclusion}, the conclusion and open issues for future work are discussed.
\section{Related Work} \label{sec:related_work}
The following section limits the discussion of related work to ensemble methods; this allows better comparability of
the CSGE compared to self-improving systems.
In \textit{ML} the term \textit{ensemble} describes the combination of multiple models.
The ensemble comprises a finite set of estimators, whose predictions are aggregated forming 
the ensemble prediction.
The theoretical justification of why ensembles can increase the overall predictive performance 
is given by the bias-variance decomposition~\cite{KW96}.
The key to ensemble methods is model diversification, i.e., how to create sufficiently different models from sample data. A comprehensive review of ensembles is given in~\cite{probRZS16}.
The most important design principles for ensembles are: 
Data, parameter, and structural diversity. Data diversity comprises ensembles trained on different subsets of the data. Well known representatives of this type are bagging~\cite{Breiman1996}, boosting~\cite{Schapire1990}, and random forest~\cite{Breiman2001}.
The idea of parameter diversity is to induce diversity into the ensemble by varying the parameters of the ensemble members.

A representative of this type is the multiple kernel learning algorithm~\cite{GA11} in which multiple kernels are combined. Lastly, structural diversity comprises the combination of different models, e.g., obtained by applying different learning algorithms or variable model types. These ensembles are also referred to as heterogeneous ensembles~\cite{MSA+12}. A well-known representative of this type is the stacking algorithm~\cite{Smyth1999}.
Another ensemble technique is Bayesian model averaging (BMA)~\cite{Bishop2006PRM} accounts for this model uncertainty when deriving parameter estimates. Hence, the ensemble estimate comprises the weighted estimate of the various model hypothesis.
Another method not to be confused with BMA is Bayesian model combination~\cite{MCS+11}. It overcomes the shortcoming of BMA to converge to a single model. 
Recently, a mixture of expert models, which comprise a gating model weighting the outputs of different submodels, gained much attention, as they determine state-of-the-art performance in language modelling~\cite{SMM+17} and multi-source machine translation~\cite{GC16}. These approaches are based on deep neural networks. Hence, they require many training samples and their weightings be barely interpretable.

In~\cite{GS16, GS18}, the \textit{CSGE} was presented in the context of renewable energy power forecasting. 
It comprises a hierarchical two-stage ensemble prediction system and weights the ensemble member's predictions based on three aspects, namely global, local, and time-dependent performance. In~\cite{GS17}, the system was extended to handle probabilistic forecasts. The approach presented in this article is a generalization of the approach to other \textit{ML} tasks.


\section{Method}
\label{sec:method}
In this section the novel \textit{Coopetitive Soft Gating Ensemble} method or short \textit{CSGE}, as proposed in \cite{GS17}, is introduced. After a brief general overview, we detail the different characteristics of the ensemble method namely soft gating, \mbox{global-,} \mbox{local-} and time-dependent-weighting. In the final sections, we give details on the \mbox{(self-)} optimization process and recommendations for training. 
\begin{figure}
  \centering
  \includegraphics[width=\columnwidth]{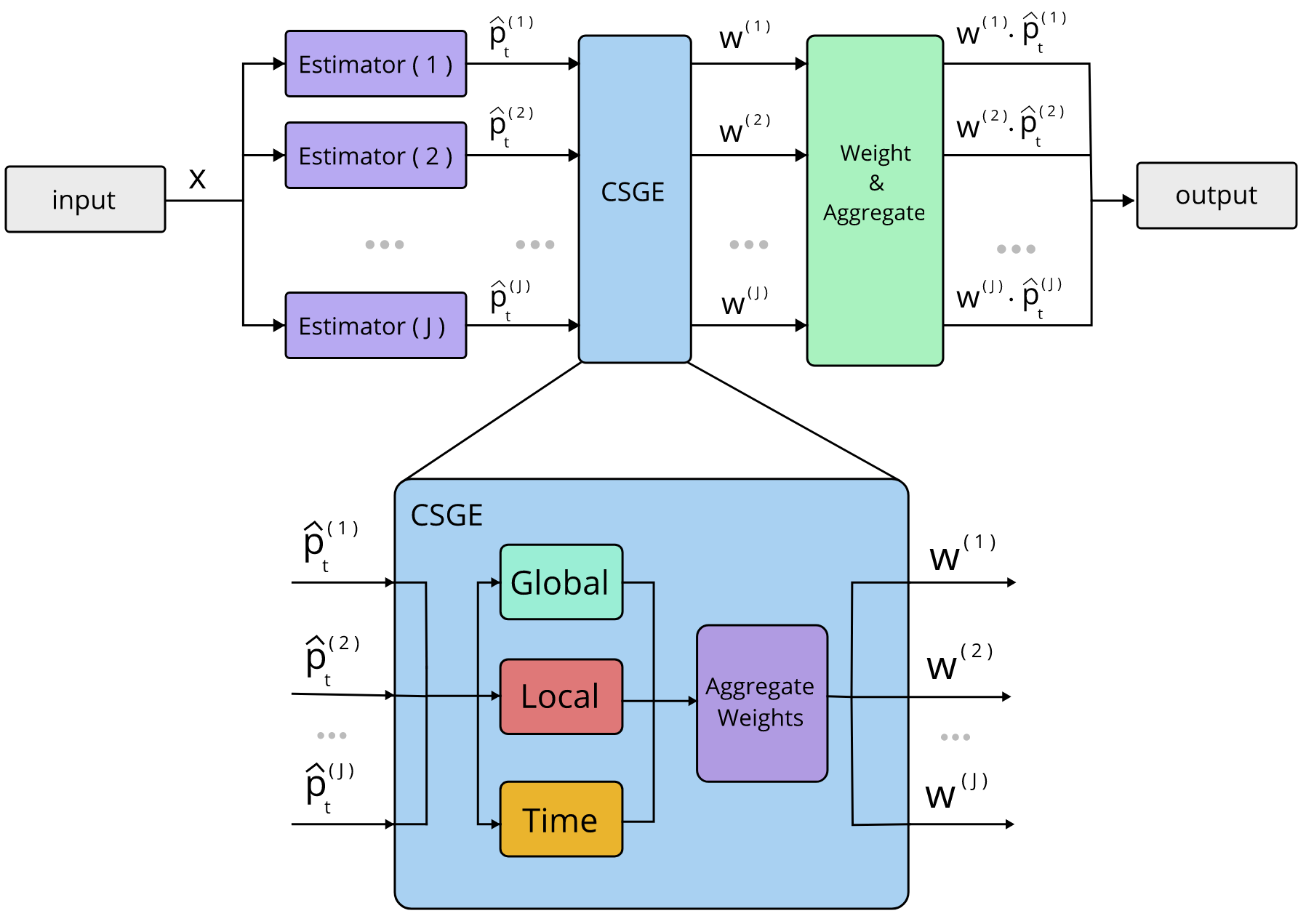}
  \vspace{0.05mm}
    \caption{The architecture of the \textit{CSGE}. The predictions $\hat{p}^{(j)}_{t}$ of the input $x$ are passed to the CSGE module. Weights are calculated regarding \textit{global-}, \textit{local-} and \textit{time-dependent weighting}. In the next step, the predictions are weighted and aggregated.\label{csge_overview_graphic}}
\end{figure}
\subsection{Coopetitive Soft Gating Ensemble}
\label{sec:csgemain}
The architecture of the \textit{CSGE}, as depicted in Fig.~\ref{csge_overview_graphic}, highlights the three weighting aspects: \mbox{global-,} \mbox{local-} and time-dependent-weighting. For each of the weighting methods the novel \textit{coopetetive soft gating} principle is applied. \textit{Coopetetive soft gating} is a conglomerate of cooperation and competetion. The ensemble combines two well known principles in ensemble methods, \textit{weighting} and \textit{gating}. \textit{Weighting} combines all ensemble members in a linear combination, while \textit{gating} selects only one of all ensemble members. The idea of the CSGE is to have the possibility to have a mixture of both \textit{weighting} and \textit{gating} and let the ensemble optimize which concept to use for the combination of different predictions.

Each of the three weighting aspects is calculate by the the predictions from $J$-ensemble members. Each ensemble member provides estimations $\hat{p}^{(j)}_{t}$ for the input $x$. $t$ denotes the timestamp $t$, also called \textit{leadtime}, when operating on timeseries for the $j$-th ensemble member. For each prediction and estimator the \textit{CSGE} calculates the local, global and time-dependent \textit{weighting} and aggregates their results. 
After normalization of $\omega^{(j)}$ each prediction $\hat{p}^{(j)}_{t}$ is weighted to obtain the final prediction as follows:


\begin{equation}\label{eq:ensemble_pred}
    \bar{\hat{p}}_{t} = \sum^{J}_{j=1} \omega^{(j)} \cdot \hat{p}^{(j)}_{t}
\end{equation}

To ensure that the prediction is not distorted weights have the following constraint:
\begin{equation}
    \label{formel_norm}
    \sum_{j=1}^{J} \omega^{(j)}=1
\end{equation}

The optimal weights $w^{(j)}$ with $j \in \{1, \dots , J \}$ are obtained by the \textit{CSGE} w.r.t. an arbitrary loss function, e.g. \textit{mean squared error}, \textit{cross-entropy} etc. Each weighting aspect has different characteristics related to the loss function summarised as follows

\begin{itemize}
\item \textbf{Global weights} are determined by observed training performance for each ensemble member and is a fixed
weighting after training. Thereby, overall strong models have more influence than weaker models. 

\item \textbf{Local weighting} considers the fact that different ensemble members have various prediction quality over the complete feature space. As an example, when considering the problem of renewable energy prediction, an ensemble member could perform well on rainy weather inputs but has worse quality when using sunny weather inputs. Therefore, the \textit{local weighting} rewards ensemble members with a higher weighting, which performed well on similar input data.
These weights are adjusted online for each prediction during runtime.

\item The \textbf{time-dependent weight} aspect is used when performing predictions on time series. Ensemble members may perform differently for different lead times. E.g., one method might achieve superior results on short time horizons, while losing quality for larger lead times. Other methods may perform worse on short time horizons, but have greater stability on larger lead times. Again, these weights are calculated online for each prediction during runtime. 
\end{itemize}

To combine these three weighting aspects for an individual ensemble member we use the  the following equation:

\begin{equation}
 \omega^{(j)} = \omega^{(j)}_g \cdot \omega^{(j)}_l \cdot \omega^{(j)}_k
\end{equation}

where $\omega^{(j)}_g$ is the \textit{global weighting}, $\omega^{(j)}_l$ is the \textit{local weighting} and $\omega^{(j)}_k$ is the \textit{time-dependent weighting}.
To calculate the final weighting the values are normalized for the $j$-th ensemble member $\omega^{(j)}$ as follows:

\begin{equation}
    \label{overallWeighingFormular}
 \omega^{(j)} = \frac{  \omega^{(j)} }{  \sum_{\tilde{j}=1}^J \omega^{(\tilde{j})} }.
 \end{equation}

This equation ensures that constraint of Eq.~\ref{formel_norm} is fulfilled.

\subsection{Soft Gating Principle}
\label{sec:sofgatingformular}


The primary goal of the \textit{CSGE} is to increase the quality of the prediction by weighting robust predictors greater than predictors with worse quality results. Traditionally in ensemble methods, one of the two paradigms \textit{weighting} or \textit{gating} are used to combine individual ensemble members. The \textit{soft gating} approach of the CSGE introduces a novel method, which allows the mixture of both \textit{weighting} and \textit{gating} and a \mbox{(self-)} optimization process to select the optimal combination of different predictions. Moreover, the soft gating approach applies to all three weighting aspects.

To evaluate the quality of an individual ensemble member, we need to relate the error of the prediction to its respective weighting. This mapping is achieved by the function $\varsigma^{'}_\eta(\Omega, \rho)$ to determine the weights of the estimator $j$ as follows:

\begin{equation}
  \varsigma^{'}_\eta(\Omega, \rho) = \frac{\sum_{j=1}^{J} \Omega_j}{\rho^{\eta} + \epsilon} , \eta \in \mathbb{R}^{+}_{0}.
\end{equation}

$\Omega$ contains reference errors of all $J$ estimators, while $\rho$ is the individual error of the estimator $j$; the user chooses parameter $\eta$. It controls the linearity of the weighting. For greater $\eta$ the \textit{CSGE} tends to work as \textit{gating}, while smaller $\eta$ results in a \textit{weighting} approach. 

\begin{figure}
  \centering
  \includegraphics[width=0.85\columnwidth]{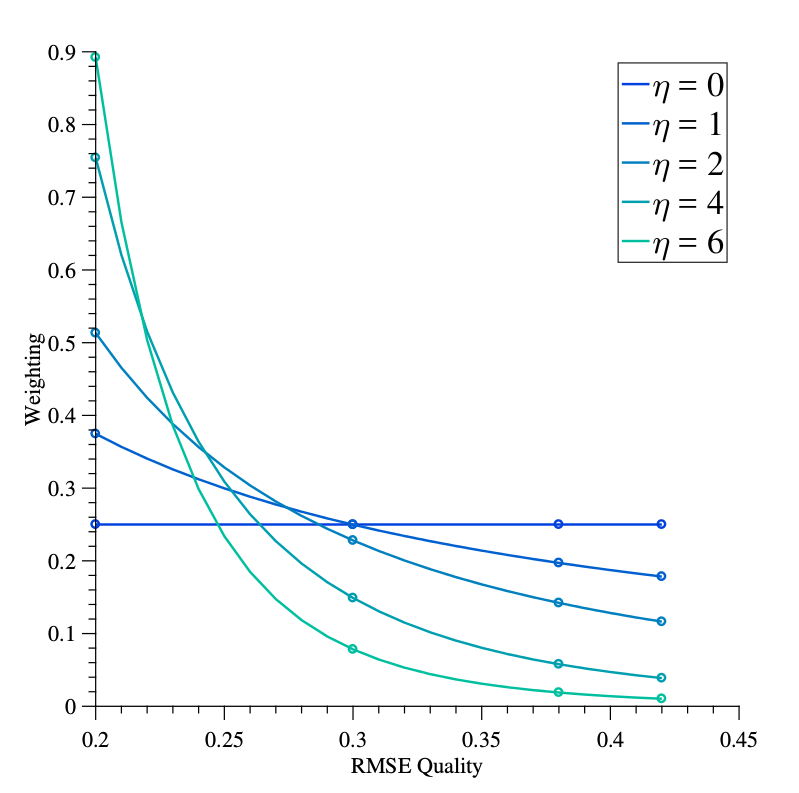}
      \caption{The error (RMSE) of a predictor is drawn on the x-axis, while the y-axis contains the corresponding weights computed by $\varsigma^{'}_\eta$. For greater $\eta$ a higher error gets more regulated with less weighting, than for smaller $\eta$.} 
      \label{etaRmseGraphic} 
\end{figure}

In Fig.~\ref{etaRmseGraphic} we observe the following characteristics of $\varsigma^{'}_\eta$:
\begin{itemize}
  \item $\varsigma^{'}_\eta$ is falling monotonously. 
  \item $\varsigma^{'}_\eta$ returns smaller weightings for an ensemble with larger errors $\rho$.
  \item For $\eta = 0$ every ensemble members are weighted with $\frac{1}{J}$, due to the later explained normalization. Respectively, disrespecting the individual errors.
  \item $\epsilon$ is a small constant to prevent a division by zero.
\end{itemize}

To ensure that  $\sum_{j=1}^{J} \omega^{(j)} = 1$, $\varsigma^{'}_\eta(\Omega, \rho)$ is adjusted in the following way

\begin{equation}
  \label{softGatingFormel}
  \varsigma^{}_\eta(\Omega, \rho) = \frac{ \varsigma^{'}_\eta(\Omega, \rho) }{ \sum_{j=1}^{J}\varsigma^{'}_\eta(\Omega, \rho_{j})}
\end{equation}

Besides the advantage on having only one parameter ($\eta$) to tune, the soft gating offers a direct relation between the weighting and the errors of the ensemble members providing a strong correlation to the actual data.

\subsection{Global Weighting}
\label{sec:global}
The \textit{global weighting} is calculated during ensemble training and then remains constant. Ensemble members that perform well on the training data get larger weights compared to those who showed a worse performance. Therefore, the difference between estimation and ground truth is calculated with

\begin{equation}
  \label{formelDiff}
  e_n^{(j)} = \theta(\hat{y}_n^{(j)},y_n)
\end{equation}

$\hat{y}_n^{(j)}$ is the prediction of the $j$-th ensemble member, while $y_n$ is the corresponding ground-truth.
$\theta$ is an a arbitrary scoring function, which could for example be the \textit{root mean-squared error (RMSE)} for regression or the \textit{accuracy score (ACC)} for classification.
The only condition is that the loss needs to be falling monotonously with increasing errors to work correctly with the \textit{soft gating principle}, see Eq.~\ref{softGatingFormel}. The error score $R^{(j)}$ of the $j$-th ensemble member is calculated by:

\begin{equation}\label{formelGlobaleGewichtungR}
  R^{(j)} = \frac{1}{N} \cdot \sum_{n=1}^{N} (e_n^{(j)})
\end{equation}

\begin{equation}
  R = ( R^{(1)}, \ldots , R^{(j)}, \ldots , R^{(J)} )
\end{equation}

By applying the \textit{soft gating} principle to the vector $R$ of all error scores of the $J$ ensemble members we obtain the final global weighting with 
\begin{equation}
  w_g^{(j)} = \varsigma^{}_\eta(R, R^{(j)})
\end{equation}

\subsection{Local Weighting}
\label{sec:local}
The \textit{local weighting} considers the quality difference between the predictors for distinct situations over the whole \textit{feature space}. Therefore, the \textit{local weighting} rewards ensemble members with a higher weighting, which performed well on similar input data. In contrast to the \textit{global weighting} the \textit{local weighting} is calculated online for each estimation during runtime.

For similar situations, we consider the distances in the input feature space. Therefore, we assume situations with low distance have more in common compared to situations with a more significant distance. $X_H$ contains all data that is used during ensemble training. Often the features of $X_H$ vary in their ranges and information value. Since we use the distances of features to determine situations which are similar, it can be useful to apply a \textit{principal component analysis} (\textit{PCA}) on the training data $X_H$.   

\begin{equation}
  X_{H_{PCA}} = PCA(X_H, N_{dim})
\end{equation}

$X_{H_{PCA}}$ is the transformed training dataset, which has a dimension of $N_{dim}$. The parameter is chosen by the user and is in the range of $1, \ldots, N_f$, where $N_f$ is the number of features of $X_H$. To calculate the local weight of a new prediction, we have to transform the input data $x$ into the transformed feature space $\hat{x}$ by applying the PCA: 

\begin{equation}
\hat{x} = applyPCA(x) 
\end{equation}

By using, e.g., \textit{k-nearest neighbor} we determine $c$ similar situations in the input data. 

\begin{equation}
  \alpha = knn(\hat{x}, X_{H_{PCA}}, c) 
\end{equation}

The vector $\alpha$ of similar situation in the input data is used to derive the errors for each situation $a$ with $e_{a}^{j}=\theta(\hat{y}_a^{(j)},y_n)$ to obtain
the average local error with:

\begin{equation}\label{knn_variante}
  q^{(j)} = \frac{1}{c} \cdot \sum_{a \in \alpha} | e_a^{(j)}|
\end{equation}

This equation is applied to each ensemble member to obtain all local error scores $Q$ for all $J$ ensemble members.

\begin{equation}
  Q = ( q^{(1)},  \ldots , q^{(j)} ,  \ldots , q^{(J)})
\end{equation}

Finally, the local weight $\omega_l^{(j)}$ is calculated by using the \textit{soft gating} principle to derive the best possible local weighting:

\begin{equation}
  \omega_l^{(j)} =  \varsigma^{}_\eta( Q , q^{(j)})
\end{equation}

\subsection{Time-Dependent Weighting}
\label{sec:time}
The \textit{time-dependent weighting} considers the fact that the quality of an ensemble member varies over leadtime. Similar to \textit{local weighting}, \textit{time-dependent weighting} is calculated for each estimation. $\hat{Y}^{(T, j)}$ contains all predictions of estimator $j$ starting at time $t=0$ to time $t=T$.

\begin{equation}
  \label{timeseriesSet}
  \hat{Y}^{(T, j)} = ( \hat{y}_{0}^{(j)},\hat{y}_{1}^{(j)},  \ldots , \hat{y}_{T}^{(j)} )
\end{equation}

The error for a specific time $t \in \{0, 1,  \ldots , T\}$ is calculated by the average error over all training samples with:

\begin{equation}
  R_t^{(j)} = \frac{1}{N} \cdot \sum_{n=1}^N e_{n}^{(t, j)} \text{~and}
\end{equation}

\begin{equation}
  e_{n}^{(t, j)} = \theta(\hat{y}_n^{(t,j)},y_n^{(t)})
\end{equation}

With $\hat{y}_n^{(t,j)} \in \hat{Y}^{(T, j)}$ and $y_n^{(t)}$ as ground truth for time $t=t$. To estimate the error score for time $t$ of estimator $j$, we use the following equation:
 
\begin{equation}
  r^{(j)}_t = \frac{R_t^{(j)}}{  \frac{1}{t + 1} \cdot \sum_{t* = 0}^{T} R_{t*}^{(j)}}
\end{equation}

$r^{(j)}_t$ is a measure that compares the error of the prediction with $t=t$ to the average error in the time intervall $t \in \{0,1,  \ldots ,T\}$. The weight $\omega_k^{(j)}$ is calculated analogous to \textit{global-} and \textit{local weighting} using the soft gating principle with

\begin{equation}
  P_t = ( r_t^{(1)} , \ldots, r_t^{(j)},  \ldots, r_t^{(J)} ),
\end{equation}

\begin{equation}
  \omega_k^{(j)} = \varsigma^{}_\eta(P_t, r_t^{(j)}),
\end{equation}
to derive the potentially best time-dependent weighting.

\subsection{Model Fusion and Ensemble Training}
\label{sec:training}

To find the optimal set of parameters for the predictions (including all weighting aspects) we aim to optimize the prediction of Eq.~\ref{eq:ensemble_pred}. Since there are three aspects, \textit{global}-, \text{local}- and \textit{time-dependent weighting}, it follows that there are also three $\eta = (\eta_0, \eta_1, \eta_2)$ to be chosen. As mentioned previously the parameter $\eta$ is chosen by the user and controls the non-linearity of the system. Therefore, the following minimization problem solves the task to adjust $\eta$ with:

\begin{equation}
\label{FormelCSGEMin}
  \sum^{N}_{n=1}[y_n - f_{CSGE}(x_n,\eta)]^2 + c \cdot \sum^{3}_{s=1}\eta_s,
\end{equation}
where $f_{CSGE}(x_n,\eta) = \bar{\hat{p}}_{t}$ is the prediction from Eq.~\ref{eq:ensemble_pred} given its current weights. $\sum^{N}_{n=1}[y_n - f_{CSGE}(x_n,\eta)]^2$ are the summed errors of the training data, while $c \cdot \sum^{3}_{s=1}\eta_s$ is a regularisation term to control overfitting.

\begin{figure}
  \centering
  \includegraphics[width=0.98\columnwidth]{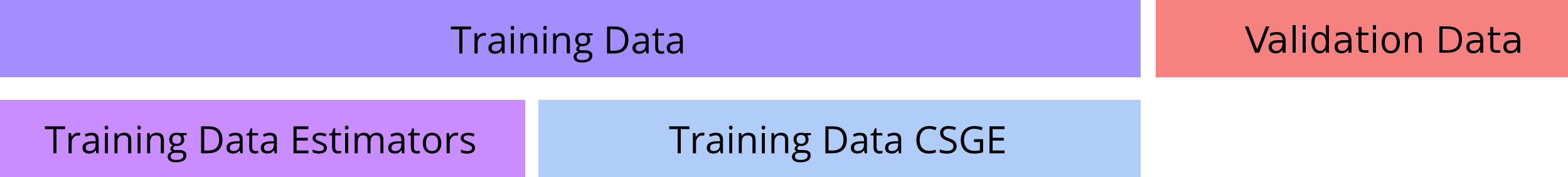}
 \vspace*{2mm}
  \caption{The two training sets must be distinct in order to get information about the quality of each ensemble member.}
  \vskip -5mm
  \label{simpleTraining}
\end{figure}

However, to optimize Eq.~\ref{FormelCSGEMin}, adjust $\eta = (\eta_0, \eta_1, \eta_2)$ and calculate the \textit{global weighting}, we need training data $E_T$. In general, the ensemble members are trained on a training dataset and validated on a validation dataset. By using the same training dataset to train the \textit{CSGE} it will often become overfitted and not generalize well. Therefore, we need training data for the \textit{CSGE} that is not used to train the $J$ ensemble members. A simple Method is shown in Fig.~\ref{simpleTraining}. The training data gets split into two sets of data. One to train the ensemble members and one to train the \textit{CSGE} itself. Even though the setup is straightforward, it has a disadvantage. The training data is wasted because the training data for the ensemble members and the one for the \textit{CSGE} need to be distinct.

\begin{figure}
  \centering
  \includegraphics[width=0.98\columnwidth]{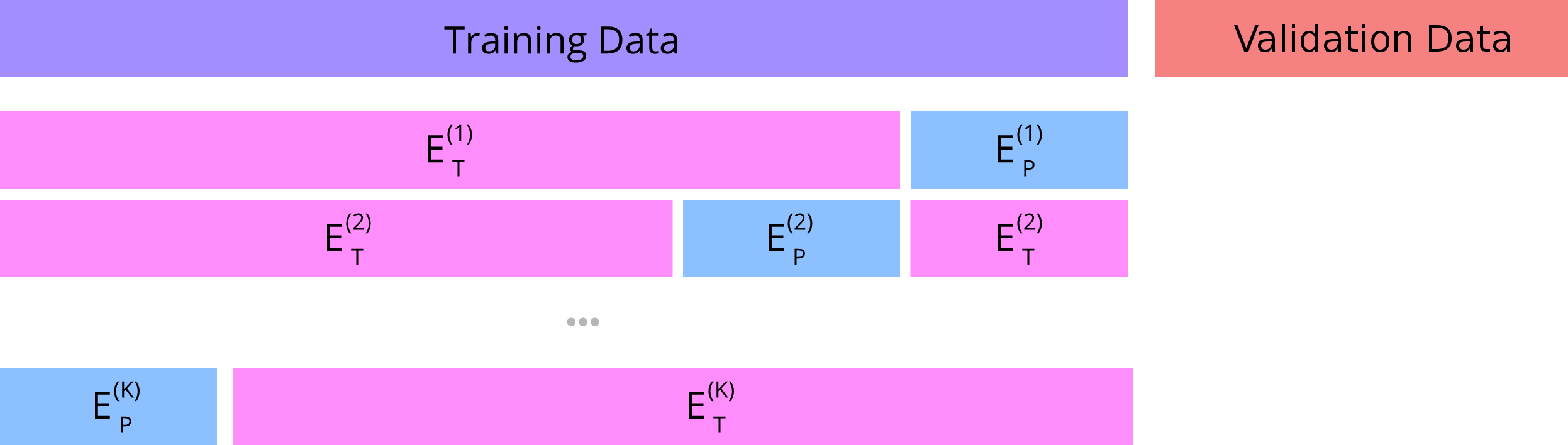}
 \vspace{2mm}
  \caption{In the $k$-th step we divide the the training data in a distinct set $E^{(k)}_T$ and $E^{(k)}_P$.  $E^{(k)}_T$ is used to train a copy of the ensemble members, while $E^{(k)}_P$ is used to make predictions. After the $k$-th iteration every element of our training data is predicted and we can use these predictions for ensemble training.}
  \label{kfoldTraining}
  \vskip -3mm
\end{figure}

A more advanced approach shown in Fig.~\ref{kfoldTraining}, allows using the training data more efficiently. Since the \textit{CSGE} uses the output data of the predictors we need those data for training. Therefore, a \textit{cross validation} with K-folds is used to generate this data. The training data in the $k$-th step of this k-fold is split in a set $E_{T}^{(k)}$ for training and a set $E_{P}^{(k)}$ for prediction. Then a copy of the $j$-th ensemble member is trained by using the set $E_{T}^{(k)}$. This temporary predictor is denoted with $f_j^{(k)}$, where $k$ is the $k$-th step and $j$ the indices for the $j$-th ensemble member. The temporary predictor $f_j^{(k)}$ is used to predict $E_{P_j}^{(k)}$, 
to concatenate all predictions in $k$-iterations.


Afterward, all $J$ ensemble members are trained on the whole training set. The training data now consists of the output data $E_{P_j}$ of the estimators; this requires us to adjust the calculation of the \textit{CSGE}. Therefore, we have to store the predictions in an $N \times J$ dimensional matrix, where $N$ is the number of samples and $J$ the number of estimators. $t$ is the timestamp when operating on time series.

\begin{equation}
C_t =
\begin{bmatrix}
C_{(t,0)}^{(0)}    & \dots     & C_{(t,0)}^{(J)}      \\
\vdots    & \ddots  & \vdots       \\
C_{(t,N)}^{(0)}     & \dots  & C_{(t,N)}^{(J)}
\end{bmatrix}
\end{equation}

Now, we have to adjust the Eq.~\ref{formelDiff}, in which the difference between prediction and ground truth is calculated. We can use $t=0$ since \textit{global-} and \textit{local weighting} do not consider the time aspect.

\begin{equation}
e_n^{(j)} = \theta(C_{(0,n)}^{(j)},y_n)
\end{equation}

Eq.~\ref{timeseriesSet}, where the set $\hat{Y}_n^{(t, j)}$ is defined, which contains all predictions of the training point $n$ of the ensemble member $j$ over the timerange $0$ to $t$.

\begin{equation}
\hat{Y}_n^{(T, j)} = ( C_{(0,n)}^{(j)},C_{(1,n)}^{(j)}, \ldots, C_{(T,n)}^{(j)} )
\end{equation}

\subsection{Regularisation Heuristic}
\label{sec:heuristic}
The ensemble learning tends to choose high $\eta$ for one single aspect and therefore $\eta=0$ for other aspects. As an example, the $\eta$ for \textit{local weighting} often are chosen very high. This example means that the local aspect of the \textit{CSGE} works as a selecting ensemble, which chooses one of the $J$ ensemble members. In order to minimise the regularisation term $c \cdot \sum^{S}_{s=1}\eta_s$, the $\eta$ of \textit{global-} and \textit{time-dependent weighting} is chosen very low. This regularization leads to an averaging ensemble for the global and time-dependent aspect, that weights all $j$ ensemble member equally with $\frac{1}{j}$, which disables these two aspects. Even though it can be necessary to disable some aspects, it is often better to distribute the values for the $\eta$'s more evenly, to get a more generalized ensemble model.
We propose the function $a(x)$ to prevent this problem. $a(x)$ weights the $\eta$ and penalises when choosing $\eta = 0$ or $\eta$ too high.
Typically the parameter $\eta$ lies in the range of $1.0 \leq \eta \leq 6.0$ \cite{GS16}.

\begin{equation}
a(x) = \frac{1}{1+e^{-\frac{1}{2} \cdot (x-10)}} + \frac{1}{2 \cdot (1+e^{x^{\frac{1}{2}}})}
\end{equation}
The minimization Eq.~\ref{FormelCSGEMin} must be adjusted in the following way
\begin{equation}
\sum^{N}_{n=1}[y_n - f_{CSGE}(x_n,\eta)]^2 + c \cdot \sum^{3}_{s=1}a(\eta_s)
\end{equation}

\begin{figure}
  \centering
  \includegraphics[width=0.9\columnwidth]{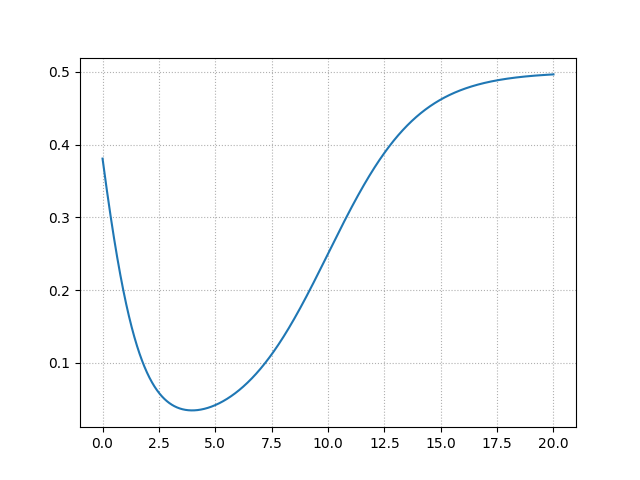}
  \caption{$a(x)$ weights the chosen $\eta$'s to avoid choosing too high or too low values for $\eta$.}
  \label{heuristic_weighting}
\end{figure}
\section{Experimental Evaluation}
\label{sec:experimente}

In this Section, we present the evaluation of the \textit{CSGE}. 
We split the evaluation into two steps. 
First, we show the proper functionality of each of the three different weighting aspects with a distinct synthetic dataset to show its interpretability aspects.
Second, we evaluate the \textit{CSGE} on four real-world reference datasets for both regression and classification. The evaluation includes a comparison with other state-of-the-art ensemble methods for a wide range of problems and shows the potential as a basic technique for SISSY systems.

\subsection{Synthetic Datasets}
\label{sec:syntheticdata}
We created synthetic datasets in order to evaluate each aspect of the \textit{CSGE}, i.e., \textit{global-}, \textit{local-} and \textit{time dependent weighting}, separately.
For each synthetic dataset, we created a data generating function $g_t$. Since we are interested in the general interpretation and functionality, we do not consider any additional noise. 
Furthermore, we defined mathematical estimators $f_j$ which have to be combined by the \textit{CSGE} 
properly to match the function $g_t$.

\subsubsection{Global Weighting}
\label{sec:evaluationglobal}
For evaluation of the \textit{global Weighting}, we created $g_t$ in the following way:
\begin{equation}
   g_t(x)=sin(x) + 4
\end{equation}

We use two estimators as ensemble members who are defined as follows:
\begin{equation*}
   f_1(x)=sin(x) \text{~and~} f_2(x)=sin(x) + 10
\end{equation*}

\begin{figure}
  \centering
  \includegraphics[width=0.9\columnwidth]{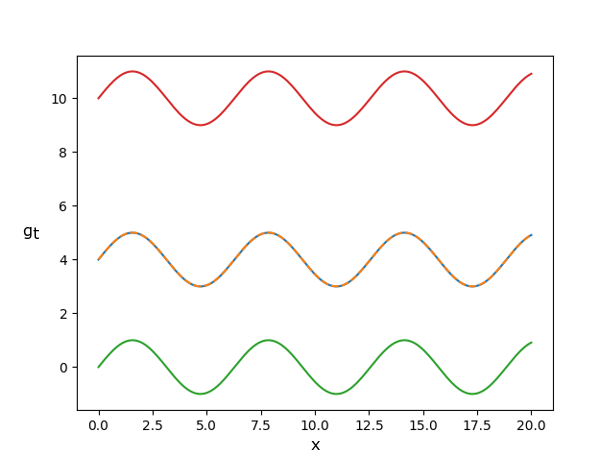}
  \caption{$g_t$ is the orange function, $f_1$ is the green function and $f_2$ is drawn in red. The predicted values by the \textit{CSGE} are drawn blue dotted (partly showing the function $f_1$, $f_2$ and $g_t$ below), which exactly match the yellow $g_t$.}
  \label{globalweighting_syn} 
\end{figure}

The result after training the \textit{CSGE} is depicted in Fig.~\ref{globalweighting_syn}.
Since there are neither local nor time-dependent aspects, the learning procedure chooses $\eta_{time}=0$ and $\eta_{local}=0$. 
When interpreting the chosen $\omega_1$ and $\omega_2$ from a mathematical point of view we observe that the chosen weights are correct, i.e., $0.6 \cdot sin(x) + 0.4 \cdot (sin(x)+10)$. The \textit{CSGE} perfectly matches $g_t(x)$.

\subsubsection{Local Weighting}
\label{sec:evaluationlocal}
For evaluation of the \textit{Local Weighting} we created $g_t$ in the following way:

\begin{equation}
   g_t(x)=\begin{cases}sin(x)& x<10 \\ sin(x) + 10 & 10  \leq x  \leq 15 \\ sin(x)& x > 15 \\ \end{cases}
\end{equation}

We use two estimators as ensemble members who are defined as follows:
\begin{equation*}
   f_1(x)=sin(x) \text{~and~} f_2(x)=sin(x) + 10
\end{equation*}

\begin{figure}
  \centering
  \includegraphics[width=0.9\columnwidth]{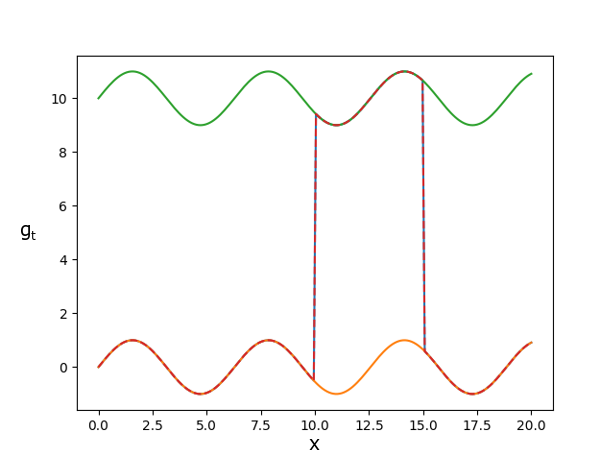}  
  \caption{$g_t$ is the red function, $f_1$ is the orange function and $f_2$ is drawn in green. The predicted values by the \textit{CSGE} are drawn blue dotted (partly showing the function $f_1$, $f_2$ and $g_t$ below), which exactly match the red $g_t$.}
  \label{localweighting_syn}
\end{figure}

This experiment has no global and time dependent aspect, therefore the learning algorithm chooses $\eta_{global} = 0$ and $\eta_{time} = 0$. Since $g_t$ can only be approximated by picking either $f_1$ or $f_2$ depending on the feature space, the chosen $\eta_{local}$ should be larger than zero. 
In Fig.~\ref{localweighting_syn}, we see the results of the experiment. We observe, that the \textit{CSGE} is able 
to perfectly reconstruct the reference model $g_t \left( x \right)$.

%


%

\subsubsection{Time-dependent Weighting}
\label{sec:evaluationtime}
For evaluation of the \textit{time-dependent Weighting} we created $g_t$ in the following way:

\begin{equation}
   g_t(x,t)=\begin{cases}sin(x)& t < 3 \\ sin(x) + 10 & t \geq 3  \end{cases}\quad t \in \mathbb{N}
\end{equation}

We use two estimators as ensemble members who are defined as follows:

\begin{equation*}
   f_1(x,t)=sin(x) \text{~and~}f_2(x,t)=sin(x) + 10
\end{equation*}

\begin{figure}
  \centering
  \includegraphics[width=0.9\columnwidth]{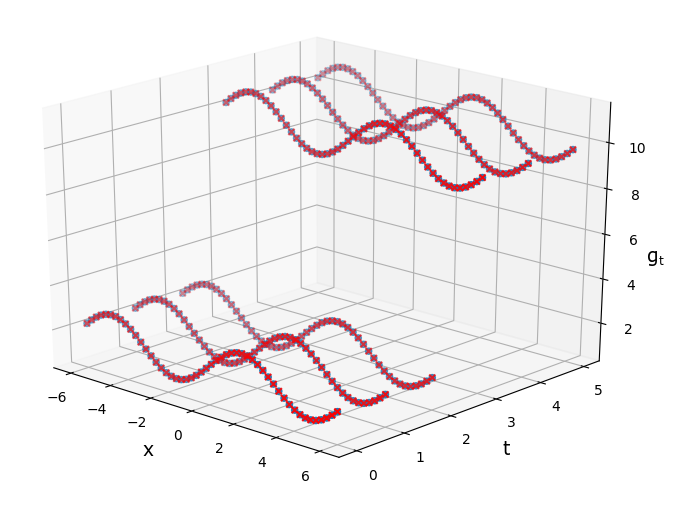}
  \caption{$g_t$ is drawn with blue points and the predictions using the \textit{CSGE} are drawn with red crosses which exactly match the $g_t$.}
  \label{timeseriesweighting_syn}
\end{figure}

Fig.~\ref{timeseriesweighting_syn} shows the results of the experiment.
Since there is no global and local aspect, the learning algorithm picks $\eta_{global} = 0$ and $\eta_{local} = 0$, $\eta_{time} > 0$. We observe that after training, \textit{CSGE} perfectly matches the reference function. These evaluations on synthetic data show that the \textit{CSGE} works properly. 
%
%



\subsection{Real-world Regression Datasets}
\label{sec:regression}
In order to evaluate the \textit{CSGE} on regression problems, we chose Boston Housing\footnote{\label{hyperref_boston_diabetes}\url{http://scikit-learn.org/stable/modules/generated/sklearn.datasets.load_[boston|diabetes].html} (last accessed: 2018/06/25)} and Diabetes datasets\footnotemark[2]. 
As ensemble members we used a \textit{Support Vector Regression} (\textit{SVR}) with radial basis function (RBF) kernel, a \textit{Neural Network Regressor} and a \textit{Decision Tree Regressor}.
As composition proceeding to the \textit{CSGE} we chose \textit{Stacking} and \textit{Voting} (i.e., \textit{Averaging}).
For \textit{Stacking} we used a \textit{Neural Network} (i.e., referred to as \textit{ANN Stacking}) and a \textit{Linear Regression} (i.e., referred to as \textit{Linear Stacking}) as meta learner. We chose the RMSE loss to optimize the \textit{CSGE}.
For each dataset, we performed ten-fold cross-validation with ten different random seeds.

We used default model parameters for the ensemble members as supplied by the scikit-learn library, 
the parameters for the ensemble methods are optimized for each experiment.
To adjust the regularisation parameter $C$ and the number of neighbors $k$ of the \textit{CSGE}, we used a grid search. 
Since the layer size of the \textit{ANN Stacking} also needs to be optimized, we applied a grid search, too.
The \textit{Linear Regression Model} has no hyper-parameters to be optimized.
As reference to \textit{CSGE} and \textit{Stacking}, we used a simple \textit{Averaging} approach.

\subsubsection{Boston Housing}
\label{sec:boston}
The overall result, i.e., RMSE, can be seen in Tbl.~\ref{boston-results}.
We observe, that both \textit{CSGE} and \textit{Stacking} achieve better results than each ensemble member. 
The \textit{Stacking} approach with a \textit{Linear Regression} meta learner achieves best results.
Even though the \textit{CSGE} has sligthly worse results compared to \textit{Linear Stacking}, it has similar performance to \textit{ANN Stacking}.

\begin{table}
\centering
\caption{RMSE on Boston Housing Dataset}
\vspace{0.5em}
\label{boston-results}
\resizebox{0.98\columnwidth}{!}{
\begin{tabular}{lcccc}
\multicolumn{5}{c}{\textbf{Ensemble Members}}                                                                              \\
                   & \multicolumn{1}{l}{}       & \multicolumn{1}{l}{}     & \multicolumn{1}{l}{}   & \multicolumn{1}{l}{} \\
\rowcolor[HTML]{C0C0C0} 
                   & \textbf{Linear Regression} & \textbf{SVM}             & \textbf{Decision Tree} & \textbf{}            \\
Mean               & 24.6673                    & 82.3656                  & \textbf{21.7313}       &                      \\
Standard Deviation & \textbf{5.8063}            & 14.7024                  & 9.9349                 &                      \\
Minimum            & 17.2139                    & 66.0964                  & \textbf{13.6710}       &                      \\
Maximum            & \textbf{33.9569}           & 107.5636                 & 46.7434                &                      \\
                   & \multicolumn{1}{l}{}       & \multicolumn{1}{l}{}     & \multicolumn{1}{l}{}   & \multicolumn{1}{l}{} \\
\multicolumn{5}{c}{\textbf{Ensemble Methods}}                                                                              \\
                   & \multicolumn{1}{l}{}       & \multicolumn{1}{l}{}     & \multicolumn{1}{l}{}   & \multicolumn{1}{l}{} \\
\rowcolor[HTML]{9B9B9B} 
                   & \textbf{CSGE}              & \textbf{Linear Stacking} & \textbf{ANN Stacking}  & \textbf{Averaging}   \\
Mean               & 18.9079                    & \textbf{15.9885}         & 18.1849                & 23.2753              \\
Standard Deviation & 8.7271                     & \textbf{5.5606}          & 5.6026                 & 8.3834               \\
Minimum            & \textbf{9.8018}            & 10.9614                  & 13.7156                & 15.7049              \\
Maximum            & 34.9028                    & \textbf{27.6670}         & 32.0342                & 39.4708              \\
\end{tabular}}
\end{table}

\subsubsection{Diabetes}
\label{sec:diabetes}

The overall result can be seen in Tbl.~\ref{diabetes-results}.
We can see, that every ensemble method achieved worse results compared to the best ensemble member (\textit{Linear Regression}).
The \textit{Linear Stacking} accomplished the best results of all ensemble methods. Nevertheless, the \textit{CSGE} performed better than \textit{Averaging} and \textit{ANN Stacking}.

\begin{table}
\centering
\caption{RMSE on Diabetes Dataset}
\vspace{0.5em}
\label{diabetes-results}
\resizebox{0.98\columnwidth}{!}{
\begin{tabular}{lcccc}
\multicolumn{5}{c}{\textbf{Ensemble Members}}                                                                              \\
                   & \multicolumn{1}{l}{}       & \multicolumn{1}{l}{}     & \multicolumn{1}{l}{}   & \multicolumn{1}{l}{} \\
\rowcolor[HTML]{C0C0C0} 
                   & \textbf{Linear Regression} & \textbf{SVM}             & \textbf{Decision Tree} & \textbf{}            \\
Mean               & \textbf{3083.1198}         & 6356.0135                & 6518.2421              &                      \\
Standard Deviation & \textbf{322.2800}          & 405.2114                 & 728.0636               &                      \\
Minimum            & \textbf{2641.9339}         & 5620.8063                & 5487.6391              &                      \\
Maximum            & \textbf{3419.9466}         & 6837.5063                & 7819.4436              &                      \\
                   & \multicolumn{1}{l}{}       & \multicolumn{1}{l}{}     & \multicolumn{1}{l}{}   & \multicolumn{1}{l}{} \\
\multicolumn{5}{c}{\textbf{Ensemble Methods}}                                                                              \\
                   & \multicolumn{1}{l}{}       & \multicolumn{1}{l}{}     & \multicolumn{1}{l}{}   & \multicolumn{1}{l}{} \\
\rowcolor[HTML]{9B9B9B} 
                   & \textbf{CSGE}              & \textbf{Linear Stacking} & \textbf{ANN Stacking}  & \textbf{Averaging}   \\
Mean               & 3333.9250                  & \textbf{3099.9531}       & 3465.7875              & 3916.1540            \\
Standard Deviation & 453.4425                   & \textbf{323.7526}        & 553.6934               & 364.5056             \\
Minimum            & 2738.1390                  & \textbf{2664.3987}       & 2795.0731              & 3380.1302            \\
Maximum            & 4273.5943                  & \textbf{3459.9381}       & 4878.9214              & 4392.5727            \\
\end{tabular}}
\end{table}

\subsection{Real-world Classification Datasets}
\label{sec:classification}

For each dataset, we performed ten-fold cross-validation with ten different random seeds.

In order to evaluate the \textit{CSGE} on classification tasks, we chose Iris\footnote{\label{hyperref_iris_wine}\url{http://scikit-learn.org/stable/modules/generated/sklearn.datasets.load_[iris|wine].html} (last accessed: 2018/06/25)} and Wine\footnotemark[3] datasets. As ensemble members we used a \textit{Support Vector Classification} (\textit{SVC}) with linear and RBF kernel and a \textit{Decision Tree Classifier}. The SVC with linear kernel is referred to as \textit{Linear Classifier}, while the SVC with RBF is referred to as \textit{SVC}. As composition proceeding to the \textit{CSGE} we chose \textit{Stacking} and majority \textit{Voting}.
For \textit{Stacking}, we used a \textit{Neuronal Network} (i.e., referred to as \textit{ANN Stacking}) and \textit{SVC} with linear kernel (i.e., referred to as \textit{Linear Stacking}) as meta learner. We chose the accuracy loss to optimize the \textit{CSGE}.

As before with the regression, we used default model parameters for the ensemble members and only 
optimized the ensemble's parameters using a grid search.
As a reference to \textit{CSGE} and \textit{Stacking}, we used the majority \textit{Voting} ensemble.

\subsection{Iris}
\label{sec:iris}

The overall results, i.e., classification accuracies, are depicted in Tbl.~\ref{iris-results}.
We observe that the \textit{CSGE} is superior to both \textit{Stacking} ensembles and \textit{Voting}. All ensemble methods results are worse than the single ensemble member, i.e., \textit{SVM}.

\begin{table}
\centering
\caption{Accuracy on Iris Dataset}
\vspace{0.5em}
\label{iris-results}
\resizebox{0.98\columnwidth}{!}{
\begin{tabular}{lcccc}
\multicolumn{5}{c}{\textbf{Ensemble Members}}                                                                              \\
                   & \multicolumn{1}{l}{}       & \multicolumn{1}{l}{}     & \multicolumn{1}{l}{}   & \multicolumn{1}{l}{} \\
\rowcolor[HTML]{C0C0C0} 
                   & \textbf{Linear Classifier} & \textbf{SVM}             & \textbf{Decision Tree} & \textbf{}            \\
Mean               & 0.6000                     & \textbf{0.9711}          & 0.9333                 &                      \\
Standard Deviation & 0.2071                     & 0.0183                   & \textbf{0.0181}        &                      \\
Minimum            & 0.2889                     & \textbf{0.9333}          & 0.9111                 &                      \\
Maximum            & 0.9778                     & \textbf{1.0000}          & 0.9556                 &                      \\
                   & \multicolumn{1}{l}{}       & \multicolumn{1}{l}{}     & \multicolumn{1}{l}{}   & \multicolumn{1}{l}{} \\
\multicolumn{5}{c}{\textbf{Ensemble Methods }}                                                                              \\
                   & \multicolumn{1}{l}{}       & \multicolumn{1}{l}{}     & \multicolumn{1}{l}{}   & \multicolumn{1}{l}{} \\
\rowcolor[HTML]{9B9B9B} 
                   & \textbf{CSGE}              & \textbf{Linear Stacking} & \textbf{ANN Stacking}  & \textbf{Voting}      \\
Mean               & \textbf{0.9578}            & 0.6756                   & 0.9378                 & 0.9511               \\
Standard Deviation & 0.0221                     & 0.1582                   & 0.0888                 & \textbf{0.0204}      \\
Minimum            & \textbf{0.9333}            & 0.4000                   & 0.6889                 & \textbf{0.9333}      \\
Maximum            & \textbf{0.9778}            & 0.9333                   & 0.9778                 & \textbf{0.9778}      \\
\end{tabular}}
\end{table}

\begin{figure}
  \centering
  \includegraphics[width=0.9\columnwidth, clip, trim = 8 0 0 0]{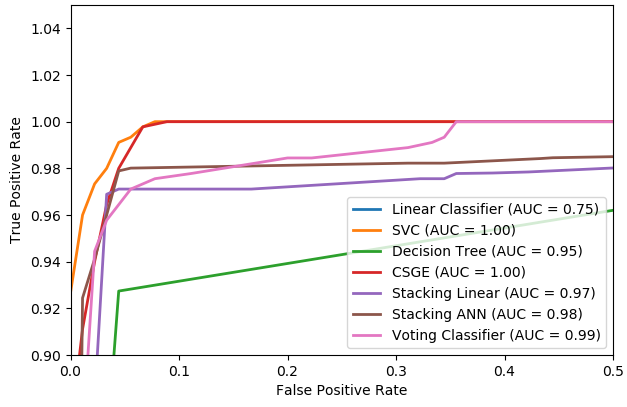}  
  \caption{ROC Curve of the different ensemble approaches and their ensemble members on the iris dataset. The Linear classifier not visible since it is below the clipping.}
  \label{roc_iris}
\end{figure}

Fig.~\ref{roc_iris} shows the ROC curve of the iris dataset, we can see that the \textit{CSGE} achieves the best results compared to \textit{Stacking} and \textit{Voting}. 

\subsection{Wine}
\label{sec:wine}
The resulting accuracies are depicted in Tbl.~\ref{wine-results}. We can see, that both \textit{CSGE} achieved the best results compared to \textit{Stacking} and \textit{Voting}.
Since the \textit{Decision Tree} is by far best ensemble member, the \textit{CSGE} worked as a gating ensemble by selecting the predictions of the \textit{Decision Tree}, only.
\begin{table}
\centering
\caption{Accuracy on Wine Dataset}
\vspace{0.5em}
\label{wine-results}
\resizebox{0.98\columnwidth}{!}{
\begin{tabular}{lcccc}
\multicolumn{5}{c}{\textbf{Ensemble Members}}                                                                              \\
                   & \multicolumn{1}{l}{}       & \multicolumn{1}{l}{}     & \multicolumn{1}{l}{}   & \multicolumn{1}{l}{} \\
\rowcolor[HTML]{C0C0C0} 
                   & \textbf{Linear Classifier} & \textbf{SVM}             & \textbf{Decision Tree} & \textbf{}            \\
Mean               & 0.4944                     & 0.4204                   & \textbf{0.9148}        &                      \\
Standard Deviation & 0.1197                     & 0.0800                   & \textbf{0.0265}        &                      \\
Minimum            & 0.3704                     & 0.3148                   & \textbf{0.8704}        &                      \\
Maximum            & 0.6852                     & 0.5185                   & \textbf{0.9444}        &                      \\
                   & \multicolumn{1}{l}{}       & \multicolumn{1}{l}{}     & \multicolumn{1}{l}{}   & \multicolumn{1}{l}{} \\
\multicolumn{5}{c}{\textbf{Ensemble Methods}}                                                                              \\
                   & \multicolumn{1}{l}{}       & \multicolumn{1}{l}{}     & \multicolumn{1}{l}{}   & \multicolumn{1}{l}{} \\
\rowcolor[HTML]{9B9B9B} 
                   & \textbf{CSGE}              & \textbf{Linear Stacking} & \textbf{ANN Stacking}  & \textbf{Voting}      \\
Mean               & \textbf{0.9148}            & 0.6907                   & 0.8759                 & 0.6704               \\
Standard Deviation & \textbf{0.0265}            & 0.1615                   & 0.0509                 & 0.1776               \\
Minimum            & \textbf{0.8704}            & 0.4444                   & 0.7593                 & 0.3704               \\
Maximum            & \textbf{0.9444}            & \textbf{0.9444}          & \textbf{0.9444}        & 0.9259               \\
\end{tabular}}

\end{table}

\begin{figure}
  \centering
  \includegraphics[width=0.9\columnwidth, clip, trim = 8 0 0 0]{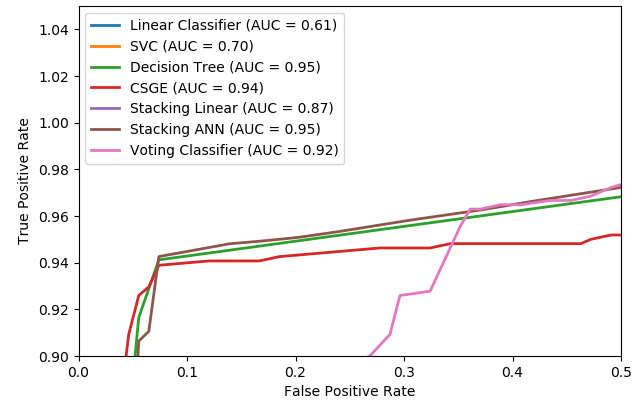}
  \caption{ROC Curve of the different ensemble approaches and their ensemble members on the wine dataset. The Linear classifier is not visible since it is below the clipping.}
  \label{roc_wine}
\end{figure}

Fig.~\ref{roc_wine} shows the ROC curve of the classifiers on the wine dataset. We observe that the \textit{CSGE} achieves the best results compared to \textit{Stacking} and \textit{Voting}. 

%
%
%

\section{Conclusion and Future Work}
\label{sec:conclusion}

In this article, we proposed the \textit{CSGE} for general machine learning tasks and interwoven systems. 
The CSGE is an ensemble method which comprises human-understandable weightings based on the three basic 
aspects as there are \mbox{\textit{global-},} \mbox{\textit{local-}} and \textit{time-dependent} weights.

The CSGE can be optimized according to arbitrary loss functions making it accessible for a broader range of problems and provides a self-improving scheme based on previously seen data. This self-improving scheme can be applied to the self-integration problem and consequently constitutes a possible basic technique for SISSY systems as outlined in~\cite{TomfordeRBW16}. Moreover, we introduced a novel hyper-parameter initialization heuristics, enhancing the training process.
We showed the applicability and easy interpretability of the approach for synthetic datasets as well as real-world data sets.
For the real-world datasets, we showed that our \textit{CSGE} approach reaches state-of-the-art performance compared to other ensembles methods for both classification and regression tasks.

For future work, we intend to apply the approach to more real-world problems in various domains, such as trajectory forecasting 
of vulnerable road users, and further investigate its applicability in other domains of \textit{AC}.

\small{
\section{Acknowledgment}
This work results from the project DeCoInt$^2$, supported by the German Research Foundation (DFG) within the priority program SPP 1835: ``Kooperativ interagierende Automobile", grant numbers SI 674/11-1. This work results from the project project Prophesy (0324104A) funded by BMWi (German Federal Ministry for Economic Affairs and Energy).
}



\bibliographystyle{IEEEtran}
%

{\small
\bibliography{IEEEabrv,references}
}
\vskip -15mm

\end{document}